\documentclass{article}
\usepackage{spconf,amsmath,amssymb,graphicx,hyperref}
\usepackage{booktabs}

\title{Pronouncing Hard Words: Phoneme Adaptation of a Pretrained\\
       Conversational Text to Speech Model}

\name{{\fontsize{10.3}{11}\selectfont
Yashish M. Siriwardena$^{1}$, Dinesh K. Gnanasekaran$^{1}$,
Tanmay Laud$^{1}$, Shinji Watanabe$^{2}$,
Subhabrata Mukherjee$^{1}$, Shanil Puri$^{1}$}}
\address{$^{1}$Hippocratic AI \qquad $^{2}$Carnegie Mellon University}

\begin{document}
\ninept
\maketitle

\begin{abstract}
Voice agents built as cascaded automatic speech recognition (ASR)
$\rightarrow$ large language model (LLM) $\rightarrow$ text-to-speech (TTS)
pipelines increasingly rely on \emph{conversational} TTS: streaming, context conditioned models that adapt delivery to dialogue history. Such models
are natural and expressive but offer little explicit control over how rare,
trust-critical drug names are pronounced, because these entities are too infrequent, and their
vocabulary too fast-growing, to be covered by any amount of pre-, mid-, or
post-training. We retrofit phoneme controllability
onto a pretrained conversational TTS model without any pre/mid-training, using an ARPAbet
vocabulary extension, continued phoneme-augmented supervised fine-tuning (SFT),
and a training-free inference-time acoustic-biasing step. On a 7{,}500-clip clinical drug-pronunciation evaluation, our model raises strict Gemini-judged correctness from $54.6\%$ (no phoneme augmentation) to $90.7\%$, outperforming every conversational and stateless TTS baseline on the automatic metric. Because no training stage sees a drug name, the entire evaluation is
out-of-domain, so the gains reflect learned pronunciation rather than memorization. In a blinded, randomized human study, adult listeners judge $84\%$ of our clips
correct, versus $82\%$ for the strongest baseline, with overlapping confidence intervals on a stratified subset. Across all 250 matched clips,
Gemini agrees with the human majority on $84.8\%$ of binary labels
($\kappa{=}0.57$).
\end{abstract}

\begin{keywords}
text-to-speech, pronunciation control, phonemes, conversational speech synthesis,
human evaluation, acoustic-biasing
\end{keywords}

\section{Introduction}
\label{sec:intro}
Recent TTS systems have moved beyond clean single-speaker narration toward
expressive, controllable, conversational generation. We target the
\emph{voice-agent} setting, where TTS is the last stage of the pipeline that turns an agent's text response into a spoken reply. That pipeline is usually a cascade; automatic speech recognition (ASR), then a large language model (LLM), then TTS, though some systems instead use a speech-to-response model that maps the user's speech directly to the agent's reply before TTS. Either way, the TTS must render each agent turn in the context of an ongoing conversation; an expert evaluation reported lower cross-turn coherence for context-agnostic XTTS than for CSM and context-conditioned Dia variants \cite{liu2025context}. This favors conversational TTS models that condition on dialogue history, including FireRedTTS-2 \cite{xie2025fireredtts2}, MOSS-TTSD
\cite{zhang2026mossttsd}, CoVoMix2 \cite{zhang2025covomix2}, and Sesame CSM \cite{sesame2024csm}. Stateless, single-utterance models such as Qwen3-TTS \cite{hu2026qwen3tts}
and CosyVoice~3 \cite{du2025cosyvoice3} instead synthesize each turn independently without any conditioning on the previously generated TTS turns or user input.

Despite this, in the conversational TTS world, explicit pronunciation control has received comparatively little
attention. This matters as voice agents are deployed in healthcare, where they meet rare, trust-critical entities
whose pronunciations are not recoverable from spelling alone. Drug names such as \textit{Mounjaro} and \textit{Xiidra} are short spans, but mispronouncing them
can dominate the perceived quality of an interaction. Mispronunciation may also create a safety risk that impedes clinical deployment.

Classical pipelines exposed pronunciation through lexicons and
grapheme-to-phoneme (G2P) conversion \cite{jurafsky2000speech}. End-to-end models simplified this but made
grapheme-to-sound decisions implicit \cite{shen2018tacotron2}. Prior work explored mixed grapheme/phoneme inputs
\cite{kastner2019representation}, phoneme-aware encoders such as PnG BERT
\cite{jia2021pngbert}, neural lexicon readers \cite{he2022nlr}, and post-hoc
correctors \cite{fong2023spell4tts}. These show explicit phonetic
information is useful, but do not answer a practical question raised by current
pretrained conversational TTS systems: can an existing large LLM based TTS model be adapted
to accept phoneme control \emph{without} any pre/mid training? This matters
because the best systems are trained at a scale most teams cannot reproduce;
FireRedTTS-2 uses 1.1M hours of monologue plus 300K hours of dialogue speech
\cite{xie2025fireredtts2}, and Qwen3-TTS reports over 5M hours across 10
languages \cite{hu2026qwen3tts}. Purpose-built phoneme-input systems such as VITS
\cite{kim2021vits}, FastSpeech2 \cite{ren2021fastspeech2}, and YourTTS
\cite{casanova2022yourtts} do accept phonemes, but they are stateless,
single-utterance models that lack the context conditioning, zero-shot cloning,
and appropriately-expressive delivery the voice-agent target requires. The problem is therefore not
``does the model accept phonemes'' but ``can pronunciation control be retrofitted
into a pretrained conversational TTS system without losing its other
capabilities?''

Extending a tokenizer with phoneme tokens is itself standard; our contributions
are the adaptation recipe, a training-free inference intervention, and a
drug-pronunciation evaluation of a conversational backbone. Concretely: (i)~a recipe to retrofit hybrid
text/ARPAbet phoneme control onto a pretrained conversational TTS model
(vocabulary extension $+$ phoneme-aware SFT), with the phoneme-annotation rate
swept across six values (0--100\%) to select an operating point;
(ii)~a training-free inference-time \emph{reference-biasing} mechanism that
conditions on a preceding-turn pronunciation reference, raising mean
drug-pronunciation accuracy by approximately $6$--$9$ points and tightening
5-run variance by approximately $6$--$9\times$ with no parameter updates, but requiring a curated recording
for each entity; (iii)~a result stronger than every tested baseline in
this curated-reference condition ($90.7\%$); and (iv)~a reference-audio-grounded pronunciation evaluation checked against a blinded, randomized
listener study ($84\%$ correct on a stratified subset).\footnote{Audio
samples, evaluation labels and Judge prompt:
\url{https://phoneme-adaptation.netlify.app/}.}\label{fn:demo}

\vspace{-0.5em}
\begin{figure*}[t]
\centering
\includegraphics[width=0.90\textwidth]{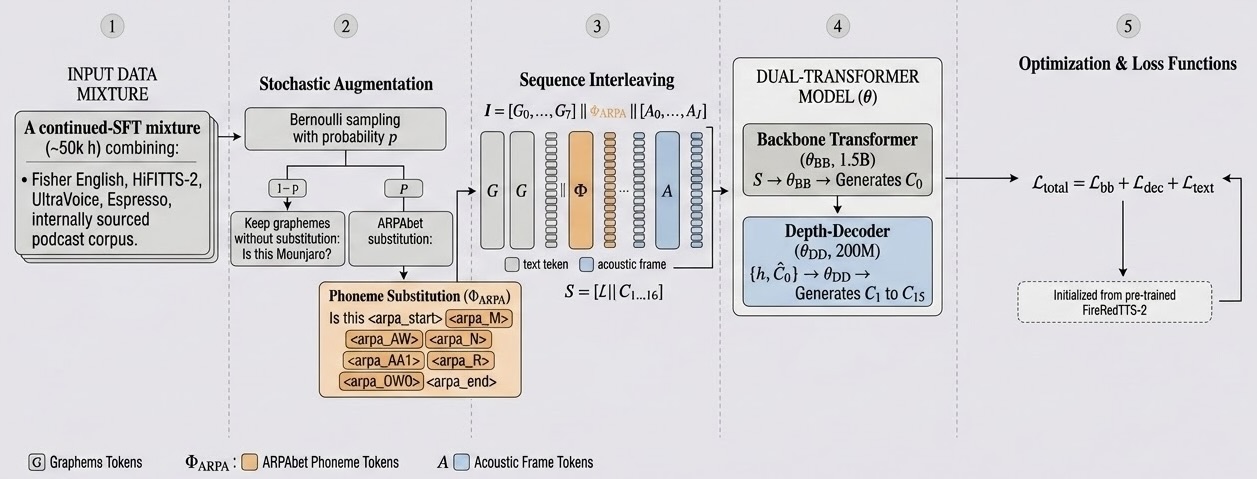}
\caption{Overview of phoneme-aware continued SFT. (1) We assemble a
multi-corpus speech mixture. (2) Stochastic augmentation retains a word's
graphemes with probability $1-p$ or replaces them with ARPAbet special tokens
with probability $p$ (illustrated for \emph{Mounjaro}; illustrative only, as no
drug name appears in SFT). (3) Grapheme or phoneme
tokens are interleaved with 16-codebook acoustic frames. (4) The
FireRedTTS-2 backbone predicts the first acoustic codebook, and its depth
decoder predicts the remaining codebooks. (5) Starting from the pretrained
checkpoint, continued SFT jointly optimizes backbone, depth-decoder, and text
losses.}
\label{fig:overview}
\vspace{-0.5em}
\end{figure*}

\vspace{-0.5em}
\section{Related Work}
\label{sec:related}

End-to-end systems \cite{shen2018tacotron2}
synthesize high-quality speech directly from text but leave few ways to specify
how rare names should be pronounced. One line retains partial phonetic control:
Representation Mixing \cite{kastner2019representation} combines character and
phoneme inputs, and \cite{fong2020limits} show limited phonemic supervision
improves controllability. A complementary line corrects pronunciation post-hoc
\cite{fong2023spell4tts}. A third enriches the text representation:
Neural Lexicon Reader \cite{he2022nlr}, PnG BERT \cite{jia2021pngbert},
Mixed-Phoneme BERT \cite{zhang2022mixedbert}, and XPhoneBERT
\cite{nguyen2023xphonebert}. Production front-ends still treat G2P, homograph
disambiguation, and post-lexical rules as active problems
\cite{comini2023multilingual}. These works mainly modify encoders, front-ends,
or correction modules, whereas we adapt the text interface of a pretrained
decoder-only conversational TTS model and leave its acoustic stack intact.

Contemporary TTS systems provide inconsistent pronunciation interfaces: some
use ARPAbet front-ends, others accept optional inline phonetic tags, and some
support IPA \cite{du2025cosyvoice3,zhou2025voxcpm}. Classical phoneme-native
models such as VITS \cite{kim2021vits}, FastSpeech2
\cite{ren2021fastspeech2}, and zero-shot YourTTS
\cite{casanova2022yourtts} accept G2P phonemes directly, but synthesize isolated
utterances without dialogue context. Our goal is therefore not to identify the
strongest phoneme-input system, but to add explicit pronunciation control to a
pretrained conversational model while retaining the capabilities needed by voice agents.

Preference- and reward-based post-training methods such as DPO and GRPO require
additional optimization; Qwen3-TTS applies DPO to human-preference pairs
\cite{hu2026qwen3tts}, while GRPO uses group-relative rewards
\cite{shao2024grpo}. We instead obtain further pronunciation gains with a training-free
inference-time intervention (Sec.~\ref{sec:biasing}). Finally, TTS quality is
usually measured on isolated utterances in clean, wideband audio with automatic
predictors (e.g., UTMOSv2 \cite{baba2024t05voicemoschallenge2024}, DNSMOS)
or standard sets (Seed-TTS
\cite{anastassiou2024seedttsfamilyhighqualityversatile}). We complement these
automatic metrics with adult-listener judgments providing an independent human evaluation.

\vspace{-0.5em}
\section{Method}
\label{sec:method}

\vspace{-0.5em}
\subsection{Base Model and Input Representation}
We start from the post-trained FireRedTTS-2 checkpoint (Apache 2.0), a long-form,
multi-speaker model built around a streaming discrete speech tokenizer and a
dual-transformer autoregressive decoder \cite{xie2025fireredtts2}: a large
decoder-only transformer (Qwen2.5-1.5B backbone) predicts the first acoustic
codebook, and a smaller depth decoder (Qwen2.5-200M) fills the remaining
codebooks, for $\approx$1.7B parameters total. Each time step is a 17-dimensional
frame: 16 acoustic-codebook coordinates plus one text-token coordinate, with the
audio end-of-segment marker represented as the all-zero frame. This factorization
gives a natural locus for adaptation: we leave the acoustic tokenizer and
decoders unchanged and modify only the text-side interface.
Figure~\ref{fig:overview} summarizes our phoneme-aware continued-SFT pipeline;
the complete recipe additionally includes inference-time reference biasing.
\vspace{-0.5em}
\subsection{Supervised Phoneme Adaptation}
\label{sec:phon-sft}
We add 131 special tokens to the tokenizer: an extended ARPAbet phone inventory
and two boundary markers, \texttt{<|arpa\_start|>} and \texttt{<|arpa\_end|>}
(full 131-token list on the demo site, Footnote~\ref{fn:demo}). The original
tokenizer has 151{,}665 text tokens while the embedding table holds 151{,}936,
so the new symbols occupy unused indices without resizing the model.

We then continue SFT from the post-trained checkpoint. Each example retains
FireRedTTS-2's interleaved text--audio format. Eligible ordinary words are
sampled from the CMU Pronouncing Dictionary \cite{cmudict} across all corpora,
not only medication-related text. An annotated word is replaced entirely by
phoneme special tokens, while the surrounding sentence remains plain text. For
example, ``Please say hello again'' may become ``Please say
\texttt{<|arpa\_start|>}\allowbreak\texttt{<|arpa\_HH|>}\allowbreak
\texttt{<|arpa\_AH0|>}\allowbreak\texttt{<|arpa\_L|>}\allowbreak
\texttt{<|arpa\_OW1|>}\allowbreak\texttt{<|arpa\_end|>}
again.'' Thus every phone, as well as the span boundaries, is a distinct
tokenizer special token. The annotation probability is the
\textit{phoneme\_tagging\_rate} $\in[0,1]$; unannotated words and examples retain
their graphemes to reduce regressions in standard synthesis.

For training, we optimize FireRedTTS-2's dual-transformer objective with added emphasis on the
new text tokens:
\begin{equation}
\mathcal{L}_{\mathrm{SFT}}
= 2\bigl((1-\lambda_{\mathrm{dec}})\mathcal{L}_{\mathrm{bb}}
 + \lambda_{\mathrm{dec}}\mathcal{L}_{\mathrm{dec}}\bigr)
+ \lambda_{\mathrm{text}}\mathcal{L}_{\mathrm{text}},
\label{eq:sft_objective}
\end{equation}
where $\mathcal{L}_{\mathrm{bb}}$ is the backbone cross-entropy over
first-codebook tokens, $\mathcal{L}_{\mathrm{dec}}$ the depth-decoder loss over
the remaining codebooks (computed on a random one-eighth frame subset
\cite{xie2025fireredtts2}), and $\mathcal{L}_{\mathrm{text}}$ a text-position
term applied to the new ARPAbet/boundary tokens so their previously unused slots
acquire useful representations. We set $\lambda_{\mathrm{dec}}=0.6$ and
$\lambda_{\mathrm{text}}=0.1$. The factor $2$ and
$\lambda_{\mathrm{dec}}=0.6$ follow the inherited FireRedTTS-2 objective;
$\lambda_{\mathrm{text}}=0.1$ strengthens gradients on the new tokens and was
not ablated.

\vspace{-0.5em}
\subsection{Inference-Time Reference Biasing}
\label{sec:biasing}
SFT lets the model read ARPAbet, but pronunciation of a specific hard entity can
still vary across independent generations. To stabilize \emph{known} difficult
entities at no training cost, we exploit the backbone's multi-speaker
conversational conditioning. Given the target ``\emph{Take your Mounjaro dose
today},'' we prepend [S1]: ``\emph{Mounjaro}'' with an isolated human recording,
then generate the distinct carrier sentence under [S2]. The reference bank
contains multiple human speakers, and [S1] and [S2] occupy separate speaker
slots. No target-utterance audio is supplied: the isolated [S1] example serves
only as prior conversational context rather than teacher forcing the [S2]
generation. This requires no parameter updates and needs only a curated bank of
reference pronunciations for the target entities
(Sec.~\ref{sec:biasing-ablation} reports the accuracy and variance gains). This
mechanism exploits a practical advantage of FireRedTTS-2's conversational
architecture: generation can condition directly on prior text-audio turns. Stateless baselines on the other hand synthesize each utterance independently and cannot accept an equivalent acoustic prior-turn prompt.

\vspace{-0.5em}
\section{Experimental Setup}
\label{sec:setup}
\vspace{-0.5em}
\noindent\textbf{SFT data.} The continued-SFT mixture ($\sim$50{,}000 h) combines
Fisher English \cite{cieri2004fisher1,cieri2005fisher2} ($\sim$1960 h) for
conversation, HiFiTTS-2 \cite{langman2025hifitts2} ($\sim$20,000 h) for clean read
speech, UltraVoice \cite{tu2025ultravoice} ($\sim$830 h) and Expresso
\cite{nguyen2023expresso} ($\sim$46 h) for expressive, paralinguistically-rich
speech, and an internally sourced conversational corpus ($\sim$28,000 h). The mixture
contains \emph{no} drug names or clinical templates, so every drug in the evaluation is out-of-domain.

\noindent\textbf{Training.} SFT initializes from the post-trained checkpoint and
runs up to 30{,}000 steps (cap 50 epochs) on 8$\times$H200, fused AdamW/bf16,
peak LR $5\!\times\!10^{-6}$, cosine decay, 500-step warmup, global batch 256.

\noindent\textbf{Automatic evaluation.} We use two evaluation sets.
\textbf{\textsc{drug\_ref}} is a 7{,}500-clip pharmaceutical-pronunciation
evaluation which covers 300 drugs $\times$ 5 templates $\times$ 5 generations. Given no
training stage sees a drug name from this hard 300 drug set, it measures generalized pronunciation rather
than memorization. \textbf{Seed-TTS}
\cite{anastassiou2024seedttsfamilyhighqualityversatile} measures general
synthesis and cloning. For drug pronunciation, a multimodal judge
(Gemini-3.1-Pro at $T{=}0$) scores each clip against a human
reference recording as \textit{correct}/\textit{partial}/\textit{wrong}; we
report the strict correct rate (mean $\pm$ 5-run $\sigma$) plus a wav2vec~2.0
entity-span phoneme error rate (PER) \cite{xu2021zeroshotphoneme}.
PER is secondary: its narrow $20$--$26\%$ range does not resolve large
differences in judged correctness (Table~\ref{tab:main_results}), so we compare
the reference-grounded judge with listeners (Sec.~\ref{sec:human-eval}).
The 300-item ARPAbet/IPA table was curated internally, and the corresponding
reference pronunciations were recorded separately from model generation. A group of linguists and clinicians validated their correctness. For FireRed reference-biasing runs, the same per-drug recording provides [S1] context and the judge reference; however, the evaluated [S2] clip has a different speaker slot and carrier sentence and is not a reconstruction of that recording.

The baseline set of TTS models used for comparison is representative rather than exhaustive. Because our target is an interactive voice agent, selection prioritized recent systems with
streaming inference or published low-latency operation, using TTFA/RTF as
practical selection criteria, alongside proprietary systems with native
pronunciation controls. We do not compare latency numerically because published
measurements use incompatible hardware, serving stacks, and metric definitions.
We also report Seed-TTS-Eval word error rate (WER) and speaker similarity (SIM)
\cite{anastassiou2024seedttsfamilyhighqualityversatile}, and UTMOSv2
\cite{baba2024t05voicemoschallenge2024} on \textsc{drug\_ref}.

\noindent\textbf{Human evaluation.} Twelve adult listeners evaluate a subset of
\textsc{drug\_ref} using the same three-point rubric as the Gemini judge:
\emph{correct}, \emph{partial}, or \emph{wrong}
(full rubric on the demo site, Footnote~\ref{fn:demo}). System identity is blinded and clip order is
randomized. The subset uses ten drugs selected by stratifying candidates across
systems and Gemini labels, with five templates per drug ($10 \times 5 = 50$ clips
per system). Each listener rates 73--75 clips, giving three or four independent
ratings per clip. A clip is human-correct with at least two of three or three of four correct ratings; exact-label ties use the lower ordinal median. We report majority-vote correct rates with 95\% Wilson CIs and
ordinal Krippendorff's $\alpha$ across the three-point ratings. We compare
Gemini with the human majority on the same clips using binary
correct/not-correct agreement, exact three-way agreement, and Cohen's $\kappa$.

\vspace{-0.5em}
\section{Results and Discussion}
\label{sec:results}

\begin{table*}[t]
\centering
\small
\begin{tabular}{lrrrrrr}
\toprule
Model & \#Params & Drug-pron $\uparrow$ & PER $\downarrow$ & UTMOSv2 $\uparrow$ & WER $\downarrow$ & SIM $\uparrow$ \\
\midrule
\textsc{phon\_0}     & 1.7B &  54.6 \,$\pm$\, 0.3 & 25.61 & 3.49 & 2.01 & 67.1 \\
\textsc{phon\_20}    & 1.7B & 86.7 \,$\pm$\, 0.7 & 22.39 & 3.48 & 1.81 & 67.2 \\
\textsc{phon\_40}    & 1.7B & 90.9 \,$\pm$\, 1.0 & 21.72 & 3.48 & 1.87 & 67.1 \\
\textsc{phon\_60} (ours) & 1.7B & 90.7 \,$\pm$\, 0.6 & 22.20 & 3.49 & \textbf{1.80} & 67.2 \\
\textsc{phon\_80}    & 1.7B & 91.5 \,$\pm$\, 0.3 & 21.41 & 3.49 & 1.92 & 66.6 \\
\textsc{phon\_100}   & 1.7B & \textbf{91.8} \,$\pm$\, 0.4 & 21.13 & 3.50 & 1.81 & 65.6 \\
\midrule
\multicolumn{7}{l}{\textit{conversational TTS baselines}} \\
FireRedTTS-2 (base)  & 1.7B & 52.3 \,$\pm$\, 0.5 & 25.81 & 3.41 & 1.95 & 66.5 \\
MOSS-TTS-Realtime \cite{gong2026mosstts} & 1.9B & 53.0 \,$\pm$\, 0.5 & 24.27 & 3.24 & 1.97 & 68.9 \\
\midrule
\multicolumn{7}{l}{\textit{stateless TTS baselines}} \\
CosyVoice3 (CMU-ARPA) & 1.5B & 82.6 \,$\pm$\, 1.2 & 25.52 & 3.62 & 2.22 & 72.0 \\
VoxCPM2              & 2B & 86.1 \,$\pm$\, 0.7 & 22.03 & 3.45 & 1.84 & \textbf{75.3} \\
ElevenLabs Turbo v2 (IPA) & --- & 73.4 \,$\pm$\, 0.5 & 20.51 & 3.46 & --- & --- \\
ElevenLabs v3 (plain) & --- & 68.0 \,$\pm$\, 0.6 & \textbf{20.03} & 3.38 & --- & --- \\
ElevenLabs v3 (IPA)  & --- & 60.9 \,$\pm$\, 0.5 & 20.59 & 3.35 & --- & --- \\
Inworld TTS-2 (IPA)  & --- & 51.5 \,$\pm$\, 1.1 & 26.06 & \textbf{3.69} & --- & --- \\
\bottomrule
\end{tabular}
\caption{Pronunciation, acoustic quality, and Seed-TTS-Eval results. Baselines are grouped into
\emph{conversational} (context-conditioned) and \emph{stateless}
(single-utterance) TTS. \textit{\#Params} is total published/inferred parameter
count (backbone + decoder where applicable; --- denotes an undisclosed
proprietary API). \textit{Drug-pron} is Gemini-judged correctness (\%),
mean $\pm$ 5-run $\sigma$; PER is Wav2Vec2 entity-span phoneme error rate (\%);
\textsc{phon\_0} and FireRedTTS-2 base receive plain graphemes; the remaining FireRed variants receive ARPAbet and
FireRed-specific reference biasing. CosyVoice3 receives CMU-ARPAbet;
ElevenLabs and Inworld rows marked IPA receive inline IPA; rows marked plain,
VoxCPM2, and MOSS receive graphemes. External systems receive no FireRed biasing.
Baseline WER/SIM values are reported Seed-TTS-Eval Test-EN results from
\cite{zhou2025voxcpm}; values for our variants use the same evaluation protocol.
Best value per column in \textbf{bold}.}
\label{tab:main_results}
\vspace{-0.5em}
\end{table*}

\vspace{-0.5em}
\subsection{SFT Unlocks Phoneme Control}
\vspace{-0.5em}
The \textsc{phon\_0} control receives the same continued
SFT mixture and compute but no phoneme augmentation, and is evaluated using
plain graphemes. It reaches $54.6\%$ \textit{Drug-pron} accuracy, only $2.3$\,pp
above the untouched FireRedTTS-2 base model ($52.3\%$), showing that continued
SFT alone explains little of the gain. At $20\%$ phoneme annotation, with
ARPAbet supplied at inference, accuracy reaches $86.7\%$, while PER moves only
from $25.61\%$ to $22.39\%$, consistent with its limited resolution. Beyond this the sweep plateaus:
\textsc{phon\_40}--\textsc{phon\_100} cluster near $91\%$ accuracy and
$21$--$22\%$ PER, so we treat the sweep as a model-selection tool rather than a
quantity to maximize blindly.

\vspace{-0.5em}
\subsection{Reference Biasing Stabilizes Known Entities}
\label{sec:biasing-ablation}
\vspace{-0.5em}
As discussed in Sec.~\ref{sec:biasing},
inference-time biasing (Table~\ref{tab:biasing_ablation}) improves every
non-trivial checkpoint by approximately $6$--$9$ points and tightens 5-run
$\sigma$ by approximately $6$--$9\times$; only FireRed rows in
Table~\ref{tab:main_results} use it. It
assumes a reference pronunciation is available, so it complements rather than
replaces phoneme adaptation; the no-bias $84.2\%$ result isolates the
phoneme-aware SFT contribution from this contextual cue. We adopt
\textbf{\textsc{phon\_60}} as our operating
point on a composite ranking over pronunciation and general quality: it has the
lowest reported Seed-TTS-Eval WER ($1.80\%$), stable UTMOSv2, and comes within
$1.1$\,pp of the top drug-pron rate; \textsc{phon\_100} is numerically best on
pronunciation but has lower SIM.

\vspace{-0.5em}
\begin{table}[t]
\centering
\small
\begin{tabular}{lrr}
\toprule
Variant   & w/ biasing $\uparrow$ & no biasing $\uparrow$ \\
\midrule
\textsc{phon\_60}  & 90.7 \,$\pm$\, 0.6 & 84.2 \,$\pm$\, 3.4 \\
\textsc{phon\_100} & 91.8 \,$\pm$\, 0.4 & 83.2 \,$\pm$\, 3.4 \\
\bottomrule
\end{tabular}
\caption{Inference-time biasing ablation (Drug-pron mean \% $\pm$ 5-run
$\sigma$).}
\label{tab:biasing_ablation}
\end{table}

\vspace{-0.5em}
\subsection{Phoneme-Input TTS Baselines}
\label{sec:phoneme-baselines}
To understand if classical G2P front-ends already solve this, we evaluated FastSpeech2/FastPitch
\cite{ren2021fastspeech2,lancucki2021fastpitch}, VITS \cite{kim2021vits}, and
YourTTS \cite{casanova2022yourtts} on \textsc{drug\_ref} under the same Gemini
protocol, both with their default G2P and with correct CMU ARPAbet injected
directly (Table~\ref{tab:phoneme_baselines}). Default G2P reaches $31$--$42\%$ and oracle phonemes reach
$39.6$--$66.3\%$, showing that accepting phoneme input alone does not guarantee correct acoustic realization of hard entities. These systems are not matched to our model in capacity or training data, so this is primarily a diagnostic comparison.

\vspace{-0.5em}
\begin{table}[t]
\centering
\small
\begin{tabular}{lrr}
\toprule
System & G2P front-end$\uparrow$ & Oracle phonemes$\uparrow$ \\
\midrule
FastSpeech2 / FastPitch  & 31.0 & 66.3 \\
VITS (LJSpeech)          & 41.7 & 55.2 \\
YourTTS                  & 32.6 & 39.6 \\
Ours (\textsc{phon\_60}) & --- & \textbf{90.7} \\
\bottomrule
\end{tabular}
\caption{Diagnostic phoneme-input comparison on \textsc{drug\_ref}
(Gemini-judged strict correct \%). \textit{Oracle}: correct CMU ARPAbet.}
\label{tab:phoneme_baselines}
\vspace{-0.5em}
\end{table}

\vspace{-0.5em}
\subsection{General TTS Quality}
\label{sec:quality}
\vspace{-0.5em}
Across the sweep,
Seed-TTS-Eval WER stays in a tight $1.80$--$2.01\%$ band, SIM stays near $67$, and
UTMOSv2 is flat (Table~\ref{tab:main_results}), indicating
that adaptation preserves these reported general-quality metrics.

\vspace{-0.5em}
\subsection{Human Evaluation}
\label{sec:human-eval}
\textbf{Adult-listener pronunciation evaluation.}
Listeners judge \textsc{phon\_60} correct on 42 of 50 clips
($84\%$, 95\% CI: $71.5$--$91.7\%$), compared with $82\%$ for VoxCPM2; its interval overlaps
those of all baselines on this 50-clip/system subset
(Table~\ref{tab:human_pron}). Pairwise exact
agreement is $79.7\%$; ordinal Krippendorff's $\alpha$ is $0.58$ (95\%
bootstrap CI: $0.49$--$0.67$), indicating moderate inter-rater reliability.
Across all 250 clips, human--Gemini agreement is $84.8\%$ for binary
correct/not-correct labels ($\kappa{=}0.57$, 95\% bootstrap CI:
$0.44$--$0.68$) and $82.4\%$ for exact three-way labels. Gemini marks $79.6\%$
correct versus $74.8\%$ by human majority; this $4.8$-point difference is not
significant (paired McNemar $p{=}0.07$). Although the $12$-point Gemini--human
gap for \textsc{phon\_60} may partly reflect its shared bias/judge reference,
the no-bias FireRed base also has an $8$-point gap and therefore we treat the
$84\%$ human rate as the conservative estimate.

\vspace{-0.5em}
\begin{table}[t]
\centering
\small
\begin{tabular}{@{}lrr@{}}
\toprule
System & Gemini (95\% CI) $\uparrow$ & Human (95\% CI) $\uparrow$ \\
\midrule
\textsc{phon\_60} (ours) & 96.0 [86.5,98.9] & \textbf{84.0} [71.5,91.7] \\
FireRedTTS-2 (base)       & 70.0 [56.2,80.9] & 62.0 [48.2,74.1] \\
MOSS-TTS-Realtime         & 68.0 [54.2,79.2] & 68.0 [54.2,79.2] \\
VoxCPM2                   & 88.0 [76.2,94.4] & 82.0 [69.2,90.2] \\
ElevenLabs Turbo v2       & 76.0 [62.6,85.7] & 78.0 [64.8,87.2] \\
\bottomrule
\end{tabular}
\caption{Adult-listener drug-pronunciation correctness (\%) on a subset of
\textsc{drug\_ref}, stratified across systems and Gemini labels ($50$ clips per
system; three or four ratings per clip). A clip is correct when at least two of
three or three of four listeners
assign \emph{correct}. Gemini scores the same clips.
Brackets give 95\% Wilson CIs.}
\label{tab:human_pron}
\vspace{-0.5em}
\end{table}

\vspace{-0.5em}
\section{Conclusion}
\label{sec:conclusion}
\vspace{-0.5em}
We present a recipe for retrofitting ARPAbet control onto a pretrained
conversational TTS backbone using vocabulary extension, phoneme-aware SFT, and
FireRed-specific reference biasing. With a curated per-drug recording bank, the
selected model reaches a result stronger than every tested baseline
($90.7\%$), versus $84.2\%$ without biasing. In a blinded, randomized human
evaluation, adult listeners judge $84\%$ of its clips correct, compared with
VoxCPM2's $82\%$ on the stratified subset, with overlapping confidence intervals.

FireRedTTS-2 is a case study rather than evidence of architecture-independent
generalization: phoneme-aware SFT may transfer to models with modifiable text
embeddings and continued-SFT support, whereas reference biasing requires
multi-turn acoustic conditioning. The recipe remains English- and ARPAbet-only,
sensitive to phoneme-annotation quality, and limited by its curated bank to
\emph{known} hard entities; transfer to other phoneme alphabets and backbones
remains future work.

\section{Compliance with Ethical Standards}
This work was supported by Hippocratic AI, which employs Y.M.S., D.K.G., T.L.,
S.M., and S.P.; the authors report no other relevant interests. Adult
listeners were employee volunteers recruited company-wide; participation was
voluntary, informed consent was obtained, and no personally identifying
information is reported or released. The study was approved by Hippocratic
AI's internal ethics committee.

\bibliographystyle{IEEEbib}
\bibliography{refs}

\end{document}